\documentclass[journal]{IEEEtran}

\ifCLASSINFOpdf
  \usepackage[pdftex]{graphicx}
\else
\fi

\usepackage[pdftex]{graphicx}
\usepackage{amsmath}
\usepackage{algorithm}
\usepackage{algpseudocode}

\usepackage{caption}
\usepackage{amsfonts}
\usepackage{array}
\usepackage{textcomp}
\usepackage{stfloats}
\usepackage{url}
\usepackage{verbatim}
\usepackage{cite}
\usepackage[dvipsnames]{xcolor}
\usepackage{soul}
\usepackage[printonlyused]{acronym}
\usepackage{comment}
\usepackage{multirow}
\usepackage{float}
\usepackage{siunitx}
\usepackage{amssymb}
\usepackage{bm}
\usepackage{nicefrac}
\usepackage{url}
\usepackage{multicol}
\usepackage{subcaption}
\usepackage{todonotes}
\usepackage{hyperref}
\hypersetup{
    colorlinks=true,
    linkcolor=blue,
    filecolor=magenta,      
    urlcolor=cyan,
    pdftitle={Overleaf Example},
    pdfpagemode=FullScreen,
    }

\newcommand{\PREPRINTYEAR}{2024}
\newcommand{\PUBLISHEDIN}{IEEE Robotics and Automation Letters}
\newcommand{\DOI}{10.1109/LRA.2024.3433749} 

\usepackage[placement=top,vshift=-7]{background}
\SetBgScale{1.0}
\SetBgContents{\PUBLISHEDIN. PREPRINT VERSION - DO NOT DISTRIBUTE. \href{https://doi.org/\DOI}{DOI \DOI}}
\SetBgColor{black}
\SetBgAngle{0}
\SetBgOpacity{1.0}

\begin{document}

\thispagestyle{empty}
\onecolumn
{
  \topskip0pt
  \vspace*{\fill}
  \centering
  \LARGE{%
    \copyright{} \PREPRINTYEAR~\PUBLISHEDIN\\\vspace{1cm}
    Personal use of this material is permitted.
    Permission from \PUBLISHEDIN~must be obtained for all other uses, in any current or future media, including reprinting or republishing this material for advertising or promotional purposes, creating new collective works, for resale or redistribution to servers or lists, or reuse of any copyrighted component of this work in other works.}
    \vspace*{\fill}
}
\NoBgThispage
\twocolumn          	
\BgThispage

\title{A Generalized Thrust Estimation and Control Approach for Multirotors Micro Aerial Vehicles}

\author{Davi Santos$^{1}$,
        Martin Saska$^{2}$,~\IEEEmembership{Member},
        Tiago~Nascimento$^{1,2}$,~\IEEEmembership{Senior Member}
\thanks{Manuscript received: April 2, 2024; Revised: June 21, 2024; Accepted: July 16, 2024. This paper was recommended for publication by Editor Giuseppe Loianno upon evaluation of the Associate Editor and Reviewers' comments. \textit{(Corresponding author: Tiago Nascimento)}}
\thanks{This work has been supported by the National Council for Scientific and Technological Development – CNPq, by the National Fund for Scientific and Technological Development – FNDCT, and by the Ministry of Science, Technology and Innovations – MCTI from Brazil under research project No. 304551/2023-6 and 407334/2022-0, by the Paraiba State Research Support Foundation - FAPESQ under research project No. 3030/2021, by CTU grant no SGS23/177/OHK3/3T/13, by the Czech Science Foundation (GAČR) under research project No. 23-07517S, and by the Europen Union under the project Robotics and advanced industrial production (reg. no. $CZ.02.01.01/00/22\_008/0004590$).}
\thanks{$^{1}$D. Santos and T. Nascimento are with the Department of Computer Systems, Universidade Federal da Paraíba, Brazil (e-mail: see http://laser.ci.ufpb.br).}
\thanks{$^{2}$T. Nascimento and M. Saska are with the Department of Cybernetics, Czech Technical University in Prague, Prague, Czech Republic (e-mail: see http://mrs.felk.cvut.cz).}
\thanks{Digital Object Identifier (DOI): see top of this page.}}

\markboth{IEEE ROBOTICS AND AUTOMATION LETTERS,PREPRINT VERSION. ACCEPTED JULY, 2024}%
{SANTOS \MakeLowercase{\textit{et al.}}: Generalized Thrust Estimation and Control for UAVs}

\maketitle

\begin{abstract}
This paper addresses the problem of thrust estimation and control for the rotors of small-sized multirotors Uncrewed Aerial Vehicles (UAVs). Accurate control of the thrust generated by each rotor during flight is one of the main challenges for robust control of quadrotors. The most common approach is to approximate the mapping of rotor speed to thrust with a simple quadratic model. This model is known to fail under non-hovering flight conditions, introducing errors into the control pipeline. One of the approaches to modeling the aerodynamics around the propellers is the Blade Element Momentum Theory (BEMT). Here, we propose a novel BEMT-based closed-loop thrust estimator and control to eliminate the laborious calibration step of finding several aerodynamic coefficients. We aim to reuse known values as a baseline and fit the thrust estimate to values closest to the real ones with a simple test bench experiment, resulting in a single scaling value. A feedforward PID thrust control was implemented for each rotor, and the methods were validated by outdoor experiments with two multirotor UAV platforms: 250mm and 500mm. A statistical analysis of the results showed that the thrust estimation and control provided better robustness under aerodynamically varying flight conditions compared to the quadratic model.
\end{abstract}


\IEEEpeerreviewmaketitle

\section{Introduction}

\begin{figure}[!t]
    \centering
    \includegraphics[width=0.5\textwidth, height=0.18\textheight]{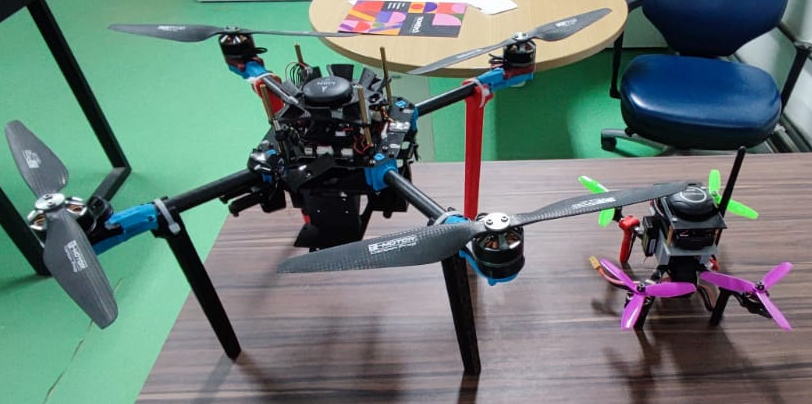}
    \caption{Quadrotors used in our experiments. The 500mm is on the left, and the 250mm is on the right.}
    \label{fig:quadrotors}
\end{figure}

\begin{figure}[!t]
    \centering
    \includegraphics[width=0.4\textwidth, height=0.11\textheight]{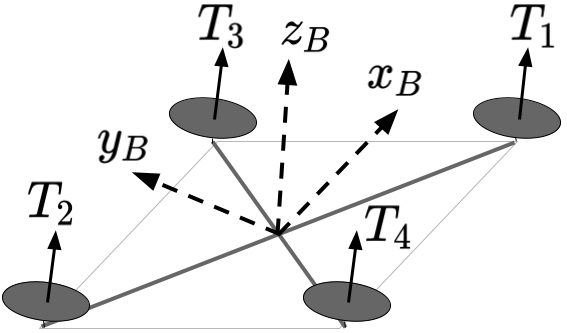}
    \caption{Quadrotors body frame and rotor thrust.}
    \label{fig:quadrotor_frames}
\end{figure}

\IEEEPARstart{M}{ultirotors} UAVs are a class of flying vehicles that have high maneuverability, making them useful for many applications such as aerial photography, search and rescue, and asset inspection~\cite{nascimento2019position}. Fig.~\ref{fig:quadrotors} presents the two UAVs used in this research, one with 500mm and the other with 250mm rotor-to-rotor distance between diagonal rotors. The rotor of this type of platform consists of a motor and a fixed-pitch propeller that rotates to produce thrust. The usual design for the quadrotor frame is in the X configuration (Fig.~\ref{fig:quadrotor_frames}), with two pairs of counter-rotating rotors that balance the torques generated by the rotation of the rotors. Thus, it is essential to control the forces applied by each rotor precisely, carefully controlling the forces on each rotor to maintain flight and achieve the desired motion, even under the influence of aerodynamic disturbances~\cite{carter2021influence}.

\begin{figure*}[htbp]
    \centering
    \includegraphics[width=\textwidth, height=0.20\textheight]{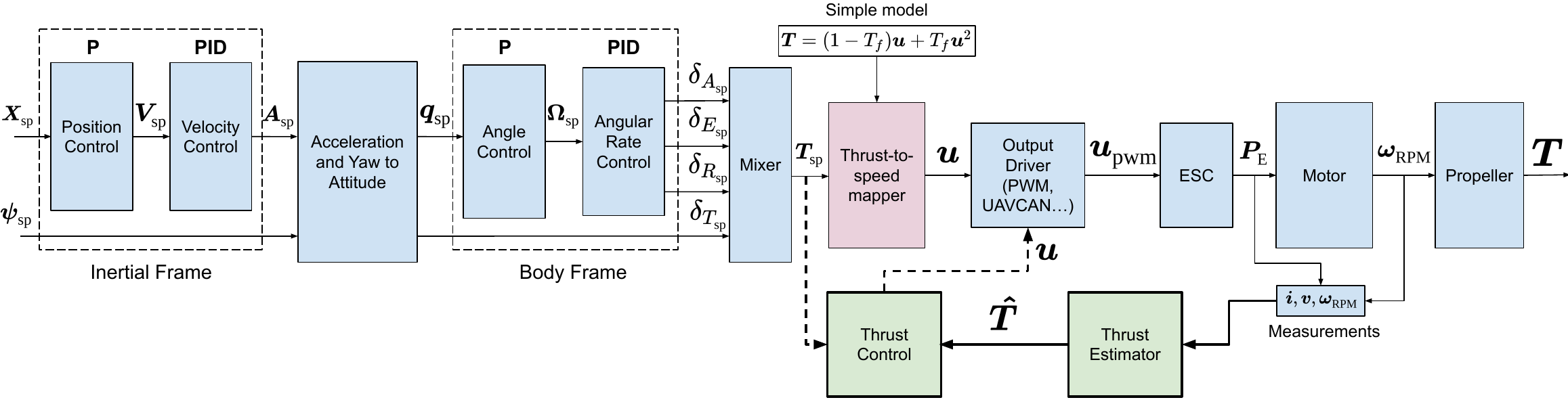}
    \caption{PX4 control pipeline. We removed the red block and added the green blocks. The thrust control and estimator blocks run at 500Hz on a Pixhawk 6C.}
    \label{fig:px4_control_pipeline}
\end{figure*}

The most common control architecture for a quadrotor consists of a cascaded hierarchical control, such as the one found on the PX4 autopilot~\cite{meier2015px4} (see Fig.~\ref{fig:px4_control_pipeline}). The PX4 environment provides a modular framework that allows for easy development, making it the most common autopilot in off-the-shelf multicopter vehicles.  In this control pipeline, the outer loop is a position controller that receives input position setpoints from higher-level navigation systems. The output of the position controller is sent to an attitude controller, which in turn outputs the desired mechanical conjugates, i.e., the forces and torques to be applied to the quadcopter frame (Fig.~\ref{fig:quadrotor_frames}). These conjugates are then sent to the control allocation matrix (e.g., the block called "Mixer"), where the desired forces and torques in the quadcopter frame are mapped to the rotor thrust, i.e., the desired thrust that, when applied by each rotor, will result in the desired forces and torques in the quadcopter frame. It is important to mention that the PX4 allows the user not to use all the controllers seen before the Mixer block, enabling the board to be coupled with any customized control architecture\cite{baca2021mrs}. In this work, we will focus on the blocks after the mixing of motors. Therefore, a mapping from rotor thrust $T$ to rotor speed $\omega$ is performed after the Mixer block. The most common model to perform this mapping is shown in equation (\ref{eq:simple_model_thrust}), where the parameters $C_{T0}$ and $C_T$ are found by static thrust experiments, being $C_T$ the thrust coefficient, and $C_{T0}$ defined as $1 - C_T$. This formulation results in a convex relationship between the angular speed of the propeller ($\omega$) and its squared value ($\omega^2$), where only one calibration parameter is needed. This mapping is suitable for hovering and low-speed flight in aerodynamic static conditions. However, this model fails in forward flight, high-speed maneuvers, and when the rotors suffer the effects of aerodynamic forces such as vertical and horizontal wind gusts~\cite{leishman2006principles}. Thus, the mapping from rotor thrust $T$ to rotor speed $\omega$ can be formulated as

\begin{equation}
    T = C_{T0}\omega + C_T\omega ^ 2  \label{eq:simple_model_thrust}.
\end{equation}

Although high-level control of multi-rotors (position, velocity, attitude, and attitude rate) is a widely researched topic~\cite{sonugur2023review,lopez2023pid,rinaldi2023comparative,khalid2023control}, thrust control or the lower levels of control for quadcopters are rarely addressed. Most authors use the standard simple thrust-to-speed map shown in equation~(\ref{eq:simple_model_thrust}). Many impressive results in quadcopter control have been achieved using this simple model, and the errors introduced by it are propagated and corrected by the complex and modern high-level control techniques. However, we argue that the most appropriate place to address this problem is at the lower levels. Thus, real-time thrust estimation and control can significantly improve high-level control performance and robustness to undesired aerodynamic effects.

\section{Related work}

The literature states that the standard approach for a quadrotor control is to use the basic quadratic thrust-to-speed map~\cite{song2021flightmare}. An improvement is to perform static thrust tests and approximate the thrust-to-speed curve with a more descriptive model. In the work of Faessler et al.~\cite{faessler2016thrust}, a polynomial equation with three coefficients to approximate the thrust mapping is used instead of the standard quadratic model. In contrast, the work of Hentzen et al.~\cite{hentzen2019disturbance} presents an approach based on the identification of the rotor thrust curve for the motor commands with the quadratic model, adding a linear term to account for the effects of depleting battery voltage. In addition, the test bench experiments performed by Kisev et al.~\cite{kivsev2022experimental} found that a cubic equation better approximated the mapping between thrust and RPM for their rotor.

Other types of work focus on describing the aerodynamic model for estimating thrust and drag in rotors from blade element theory (BET) and blade element momentum theory (BEMT), based only on the propeller parameters~\cite{gill2017propeller}. However, such approaches need to be validated by several wind tunnel experiments. In contrast, other approaches consider the whole rotor system, both the propeller and the motor dynamics, using the electromechanical feedback of voltage, current, and speed to estimate the power balance in the rotor~\cite{bangura2017thrust,Sarah_2021,papadimitriou2022external}.

A recent application driving research into more precise rotor thrust control is drone racing. In the extreme situations of high-speed maneuvering required for this application, it becomes clear that a simple quadratic thrust-to-speed map from static test curves is not efficient because these basic model assumptions neglect other significant aerodynamic effects that occur during forward flight, such as linear rotor drag, dynamic lift, rotor-to-rotor, and rotor-to-body interactions, and aerodynamic body drag~\cite{hanover2023autonomous}. A more recent successful approach combines BEMT-based models with data-driven approaches to achieve high performance~\cite{bauersfeld2021neurobem,sun2022comparative}. The problem with these approaches is that they are time-consuming, and the models developed are tuned for a specific platform in specific applications. Our approach has a broader range of applications and is easy to use on many platforms, balancing model accuracy with computational performance.

Therefore, this paper proposes a closed-loop thrust estimation and control to replace the simple thrust-to-speed map~(see equation~(\ref{eq:simple_model_thrust})) commonly used in current multirotor autopilots. This thrust control allows the low-level controller to achieve the desired thrust set better by the high-level position controller, reducing aerodynamic effects and improving flight robustness. The thrust estimates for each rotor are fed back to the control using a state-of-the-art thrust estimator. We have developed a methodology that allows the use of this thrust estimation in a wide range of multicopters with minimal calibration. Thus, the contributions of this work can be described as:
\begin{itemize}
    \item The novel generalized thrust estimation methodology for use in multiple types of flying robots;
    \item The proposal of a generalized thrust control law;
    \item The implementation of the proposed algorithms in PX4, validating the results in two different platforms and demonstrating its general use.
\end{itemize}

\section{Aerodynamic power modeling}
\label{sec:thrust_estimation_1}

To perform thrust control, it is necessary to observe or estimate the thrust at each rotor. The thrust estimation presented here is based on the aerodynamic models given by Blade Element and Momentum Theory (BEMT) for quadrotors with fixed-pitch blades, using electromechanical power measurements for speed, current, and voltage provided by each rotor ESC. The complete analysis of BEMT-based thrust estimation is explored in more detail in \cite{gill2017propeller,bangura2017aerodynamics} and summarized here.

\begin{figure}[htbp]
    \centering
    \includegraphics[width=0.3\textwidth, height=0.15\textheight]{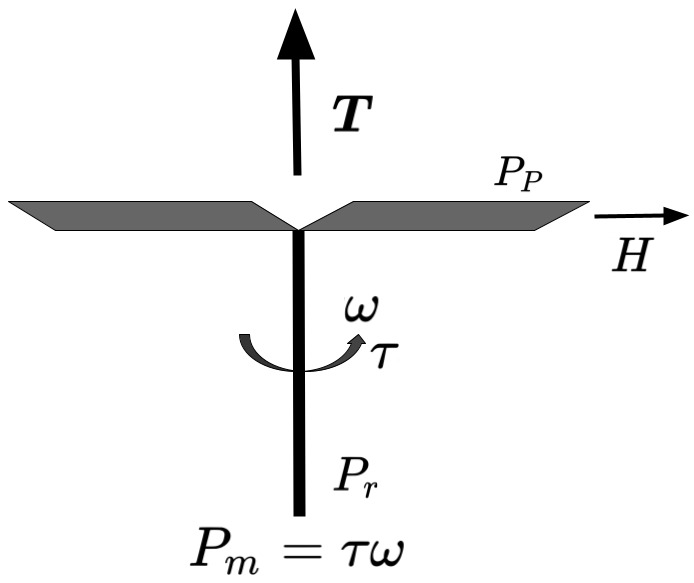}
    \caption{Different forces, torques, and powers on a rotor.}
    \label{fig:power_in_rotor}
\end{figure}

The aerodynamics of a rotor blade can be modeled by the different powers shown in Fig.~\ref{fig:power_in_rotor}. The mechanical power of the motor ($P_m$) is transmitted to the blades, resulting in a thrust power ($P_T$). There is also the horizontal power on the blade ($P_H$), which is caused by the incoming relative air due to the forward motion, and the blade profile power ($P_p$) due to the shape and pitch of the blades. A dissipative power due to the acceleration/deceleration of the rotor is represented by the friction power in the shaft ($P_r$), given by

\begin{equation}
    P_r = I_r\omega\dot{\omega},
    \label{eq:shaft_power}
\end{equation}
\noindent where $I_r$ is the moment of inertia of the rotor and $\omega$ is the rotor speed in RPM. Furthermore, the mechanical power is given as

\begin{equation}
    P_m = \tau \omega = K_q i_a \omega,
    \label{eq:mechanical_power}
\end{equation}
\noindent where $\tau$ is the torque from the propeller's rotation, $K_q$ is the motor coefficient (with $K_q=K_{q_0}-K_{q_1}\dot i_a$, where $K_{q_0}$ and $K_{q_1}$ are motor coefficients found through experiments), and $i_a$ is the current on the motor.

Through a power balance analysis on the rotor, the aerodynamic power ($P_{\text{am}}$) can be calculated as

\begin{subequations}
\begin{align}
P_{\text{am}} &= P_m - P_r \\
 &= K_qi_a\omega - I_r\omega\dot{\omega}.
 \label{eq:pam}
\end{align}
\end{subequations}

In contrast, we can also find the model for the coefficient of the aerodynamic mechanical power ($C_{P_{\text{am}}}$) and thrust ($T$) for hovering and aerodynamic static situations, with equations~(\ref{eq:pam_simple}) and (\ref{eq:pam_simpleT}).

\begin{subequations}
\begin{align}
     C_{P_{\text{am}}} = \frac{P_{\text{am}}}{\omega^3},\label{eq:pam_simple}\\
    T = C_{T}\omega^2.
    \label{eq:pam_simpleT}
\end{align}
\end{subequations}

Furthermore, from the momentum theory, one can calculate the coefficients of thrust ($C_T$), horizontal force ($C_H$), and aerodynamic power ($C_{P_{a}}$) as

\begin{subequations}
\begin{align}
    C_T &= 2\rho A R^2\lambda^i\sqrt{\mu + \lambda^2},\label{eq:ct_mt1} \\
    C_H &= 2\rho A R^2\mu^i\sqrt{\mu^2 + \lambda^2},\label{eq:ct_mt2}\\
    C_{P_{a}} &= \left(\kappa C_T\lambda^i + C_T\lambda^s + \kappa C_H\mu^i + C_H\mu^s\right)R,\label{eq:ct_mt3}
\end{align}
\end{subequations}

\noindent where $A$ is the area of the rotor disc, $\rho$ is the density of air, $\lambda$ is the vertical inflow ratio given by $\lambda = \lambda^i + \lambda^s$, where $\lambda^i$ is the induced inflow and $\lambda^s$ is the stream inflow, and $\mu$ is the advance ratio.

Finally, $\kappa$ is the induced power factor, which accounts for the additional power/energy dissipated due to wake rotation, tip loss, and non-uniform flow and is given by

\begin{equation}
    \kappa = d_0 + d_1 C_T,
    \label{eq:kappa}
\end{equation}
\noindent where $d_0$ and $d_1$ are constants.

In contrast, from blade element theory, the same coefficients can also be found with

\begin{subequations}
\begin{align}
    C_T &= \frac{1}{4} N_b \rho c_{tip} R^3 C_{l\alpha} (\theta_{tip}(2 + \mu^2) - 2\lambda) \label{eq:ct_bemt1}, \\
    C_H &= \frac{1}{2} N_b \rho c_{tip} R^3 \mu \left[C_{d0} + \frac{1}{2}X\right] \label{eq:ch_bemt2},   \\
    C_{P_{\text{am}}} &= \frac{1}{4} \rho N_b c_{tip} R^4 C_{d0} (2 + 5\mu^2) + \label{eq:cpam_bemt1}\\ &(C_T (\kappa \lambda^i + \lambda^s) + C_H(\kappa \mu^i + \mu^s)) R \label{eq:cpam_bemt},
\end{align}
\end{subequations}

\noindent where $X$ is given by

\begin{footnotesize}
\begin{equation}
    X = C_{l\alpha} \left(\theta_{tip} (\lambda + b_1 + \lambda a_0) + 2\lambda \left(\frac{4\theta_{tip}}{3} - \lambda\right) - 2\lambda b_1\right),
\end{equation}    
\end{footnotesize}

\noindent $N_b$ is the number of blades, $c_{tip}$ is the tip chord, and $\theta_{tip}$ is the tip pitch, $C_{l\alpha}$ and $C_{d0}$ are the lift and drag coefficients of the propeller and $a_0, a_1$ and $b_0$ are the flapping coefficients for a non-flapping rotor defined by Brawnwell et. al.~\cite{bramwell2001bramwell}.

\section{A Generalized thrust estimation and control approach}
\label{sec:thrust_estimation_2}

By modeling the variation of $C_T$ and $C_{P_{\text{am}}}$ during flight and then measuring $P_{\text{am}}$ and $\omega$ from the ESC telemetry feedback, it is possible to reconstruct the aerodynamic condition of the rotor, compute $C_T$ and consequently compute the actual thrust $T$ from equation~(\ref{eq:pam_simpleT}). Furthermore, the above-mentioned aerodynamic power modeling resulted in seven parameters $(C_T, C_H, \mu^i, \mu^s, \lambda^i, \lambda^s, \kappa)$ and four constraint equations $(\ref{eq:kappa}),(\ref{eq:ct_bemt1}), (\ref{eq:ch_bemt2}$), and $(\ref{eq:cpam_bemt})$. Within this modeling, $P_{\text{am}}$ and consequently $C_{P_{\text{am}}}$ can be estimated from the electrical measurements given by the ESC through equation~(\ref{eq:pam}).

There is a known relationship between the horizontal velocities and the force $H$, and thus, also a relationship between $\mu^s$, $\mu^i$, and $C_H$, with the horizontal acceleration of the vehicle. In the work of Bangura and Mahony \cite{bangura2017thrust}, these three parameters were found analytically since their estimator was implemented at the ESC level. Furthermore, we assume that the advance ratio $\mu$ is small and that $\mu^2 \approx 0$ \cite{bangura2017thrust}, resulting in the simplification of some of the equations (i.e., equations (\ref{eq:ct_mt2}), (\ref{eq:ct_bemt1}), and (\ref{eq:cpam_bemt1})). Then, by assuming that horizontal speeds are small and that the term $C_H(\kappa\mu^i + \mu^s) \propto \mu^2$, we can decouple the equation~(\ref{eq:ch_bemt2}) resulting in four aerodynamic parameters $C_T$, $\lambda^s$, $\lambda^i$ and $\kappa$, and four equations (\ref{eq:ct_mt1}, \ref{eq:kappa}, \ref{eq:ct_bemt1} and \ref{eq:cpam_bemt1}). Then, we can express the remaining equations in coefficients, defining them as

\begin{subequations}
\begin{align}
    c_0 &= R, \quad c_1 = \frac{1}{2} N_b \rho c_{tip} R^3 C_{l\alpha}, \\
    c_2 &= \theta_{tip}, \quad c_3 = \frac{1}{2}\rho c_{tip} N_b C_{d0}R^4.
\end{align}
\end{subequations}

Finally, equations~(\ref{eq:ct_mt1}), (\ref{eq:ct_bemt1}), and (\ref{eq:cpam_bemt1}) can be re-written as

\begin{subequations}
\begin{align}
    C_T &= c_1 [c_2 - \lambda], \label{eq:ct_bemt_coef} \\
    C_{P_{\text{am}}} &= c_3 + C_T (\kappa\lambda^i + \lambda^s) c_0, \label{eq:cpam_coef} \\
    C_T &= c_4\lambda^i\lambda,
    \label{eq:ct_mt_coef}
\end{align}
\end{subequations}

\noindent where equation~(\ref{eq:ct_mt_coef}) derives from equation~(\ref{eq:ct_mt1}), and~$c_4 = 2\rho A R^2 = 2\rho A c_0^2$. 

Now, combining equations~(\ref{eq:ct_bemt_coef}) and (\ref{eq:ct_mt_coef}), we get

\begin{equation}
    c_4 (\lambda^i)^2 + \lambda^i (c_4 \lambda^s + c_1) + c_1 (\lambda^s - c_2) = 0.
    \label{eq:c_4}
\end{equation}

In summary, the aerodynamic parameters $C_T$, $\lambda^s$, $\lambda^i$, and $\kappa$ can be found by the constraint equations~(\ref{eq:kappa}), (\ref{eq:ct_bemt_coef}), (\ref{eq:cpam_coef}), and (\ref{eq:ct_mt_coef}), which in turn depend on the aerodynamic coefficients $d_0, d_1, c_0, c_1, c_2, c_3$, and $c_4$. These coefficients are determined offline through laboratory experiments.

To solve these equations, an iterative approach is proposed, noticing that once $\lambda^s$ is found, it is straightforward to solve the equations and compute $C_T$, $\kappa$, and $C_{P_{\text{am}}}$. This approach first guesses a value of $\lambda^s$ and then uses the equations to calculate the aerodynamic power coefficient $C_{P_{\text{am}}}$ based on the value of $\lambda^s$. Then, we compare the result to the current value of $C_{P_{\text{am}}}(t)$, calculated based on the actual electrical measurements given by the ESC, as in $f(\lambda^s) = C_{P_{\text{am}}}(t) - C_{P_{\text{am}}}$. The goal is to find $\lambda^s$ such that it makes $f(\lambda^s) \rightarrow 0$. Afterward, we use $\lambda^s$ to solve the equation~(\ref{eq:c_4}) and find $\lambda^i$. We also use equation~(\ref{eq:ct_bemt_coef}) to estimate $C_T$ and equation~(\ref{eq:kappa}) to find $\kappa$, which in turn is used to compute $C_{P_{\text{am}}}$ from equation~(\ref{eq:cpam_coef}). Finally, $\lambda^s$ is updated at each iteration using the Newton-Secant approach, and this process is repeated N times until $\lambda^s$ converge. We summarized this process in Algorithm~$1$ in the same manner as Bangura and Mahony \cite{bangura2017thrust}.

\begin{algorithm}
\caption{Thrust Estimation.}
\begin{algorithmic}[1]
\label{alg:thrust_estimation_algorithm}

\State Data $c_0, c_1, c_2, c_3, c_4, d_0, d_1, N, \Delta, \epsilon$
\State Local state old[$\lambda^s_k$]
\State For each measurement $C_{P_{\text{\text{am}}}}(t) = \frac{\hat{P}_{\text{am}}}{\omega^3}$ at time t.
\State Set $k = 0$; Set $\lambda^s_0 =$ old$[\lambda^s_k] - \Delta$
\State \textbf{for} $k = 0 \dots N-1$ \textbf{do}
    \State \hspace{\algorithmicindent} \textbf{if} $k = 1$ \textbf{then}; Set $\lambda^s_1$ = old[$\lambda^s_k$];
    \State \hspace{\algorithmicindent} Use (\ref{eq:c_4}) to compute $\lambda^i$;
    \State \hspace{\algorithmicindent} Use (\ref{eq:ct_bemt_coef}) to compute $C_T$;
    \State \hspace{\algorithmicindent} Use (\ref{eq:kappa}) to compute $\kappa$;
    \State \hspace{\algorithmicindent} Use (\ref{eq:cpam_coef}) and (\ref{eq:c_4}) to compute $C_{P_{\text{am}}}$;
    \State \hspace{\algorithmicindent} Compute $f(\lambda^s_k) = C_{P_{\text{am}}}(t) - C_{P_{\text{am}}}$;
    \State \hspace{\algorithmicindent} \textbf{if} $k > 1$ \textbf{and} $|f(\lambda^s_k) - f(\lambda^s_{k-1})| < \epsilon$ \textbf{then break};
    \State \hspace{\algorithmicindent} Compute $\lambda^s_{k+1} = \lambda^s_k - f(\lambda^s_k) \frac{\lambda^s_k - \lambda^s_{k-1}}{f(\lambda^s_k) - f(\lambda^s_{k-1})}$; \textbf{return}
\State Set old$[\lambda^s_k] = \lambda^s_k$;
\State Output $T = C_T \omega^2$;

\end{algorithmic}
\end{algorithm}

\subsection{Generalizing the thrust estimation}

The thrust estimation algorithm presented in Bangura and Mahony \cite{bangura2017thrust} depends on the identification of several aerodynamic parameters ($c_{0\mbox{--}4}, d_{0\mbox{--}1}, K_{q_{0\mbox{--}1}}$). Usually, these parameters are found through exhaustive and time-consuming laboratory experiments, using setups to emulate the aerodynamic conditions near the rotor during forward flight~\cite{bangura2016aerodynamics}. For example, Bangura and Mahony~\cite{bangura2017thrust} identified the aerodynamic parameters for a propulsion system consisting of a Hobbyking D2836/9 950KV BLDC motor and a 10-in Hobbyking propeller with dual blades. These parameters are shown in Table~\ref{tab:aerodynamic_coefficients}.

\begin{table}[htbp]
\centering
\caption{Aerodynamic Coefficients}
\begin{tabular}{|c|c|}
\hline
\textbf{Coefficient} & \textbf{Value}                           \\  \hline
$d_0$        & $4.2959$                       \\  \hline
$d_1$        & $-1.7154 \times 10^5$         \\  \hline   
$c_0$        & $0.0724$                       \\  \hline
$c_1$        & $6.1490 \times 10^{-5}$        \\  \hline
$c_2$        & $0.2993$                       \\  \hline
$c_3$        & $1.2998 \times 10^{-8}$        \\  \hline
$c_4$        & $2\rho A c_0^2$                            \\  \hline
$K_{q0}$        & $0.242$                     \\  \hline
$K_{q1}$        & $0.0014$                    \\  \hline
\end{tabular}
\label{tab:aerodynamic_coefficients}
\end{table}

The works found in the literature usually propose approaches for thrust estimation, applying them to a specific type of robot. Suppose an estimator is needed in a different type of UAV. In that case, all the laborious calibrations required to find suitable aerodynamic coefficients for that platform must be repeated. It is also advisable to recalibrate the aerodynamic parameters for different motors of the same model. Thus, a new calibration must also be performed for the four rotors of each quadrotor. This is not feasible or desirable for most applications. Our proposed approach eliminates this process and enables thrust control for a wide range of quadrotors. Thus, we implemented the proposed method at the PX4 software architecture level to achieve wide usability.


\begin{figure}[htbp]
    \centering
    \includegraphics[scale=0.045]{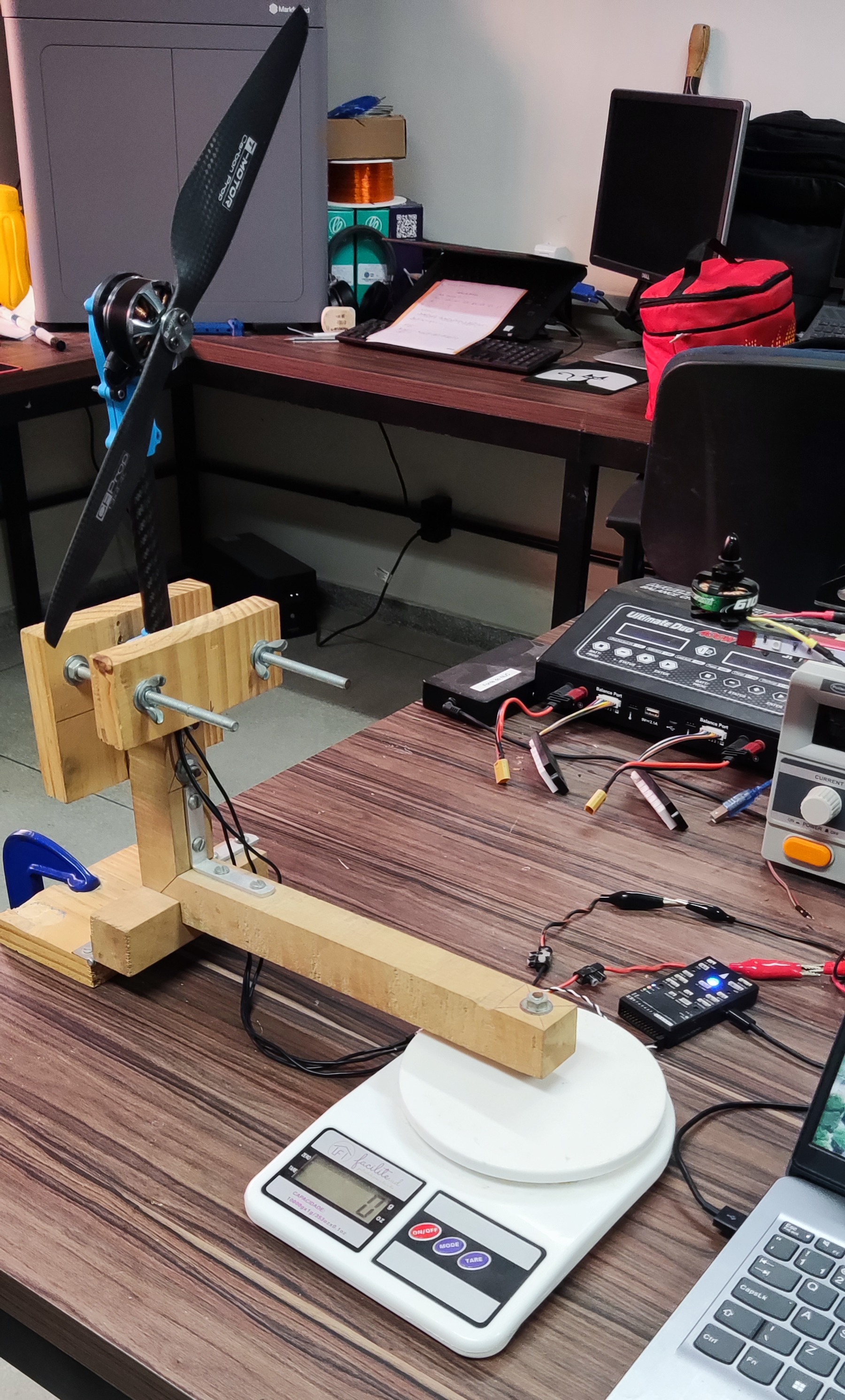}
    \caption{Setup used for the thrust test bench experiment.}
    \label{fig:experiment_setup}
\end{figure}

We started our methodology firstly by using the parameters in Table~\ref{tab:aerodynamic_coefficients} as a baseline. The observed raw thrust estimation values are acquired from performed laboratory experiments that use an L-shaped apparatus with vertical arm length L and horizontal arm length H (see Fig.~\ref{fig:experiment_setup}). The motor is placed at the top of the vertical arm, and a precision scale is used at the end of the horizontal arm to measure the thrust on the scale according to

\begin{equation}
    T = \left( \frac{L}{H} \right) F_{s},
\end{equation}

\noindent where $F_s$ is the force measured by the scale. 

\begin{table}[]
\centering
\caption{The specifications for the 250mm and 500mm quadrotors.}
\begin{tabular}{|c|c|c|}
\hline
\textbf{Component}        & \textbf{250mm}         & \textbf{500mm}           \\ \hline
Flight Controller Board & Pixhawk 6C    & Pixhawk 6C      \\ \hline
Firmware        & PX4           & PX4             \\ \hline
GPS              & M9N           & M9N             \\ \hline
Motor model      & DSMX MT2204   & T-Motor MN3510  \\ \hline
Motor KV         & 2300KV        & 700KV           \\ \hline
ESC              & 51A DSHOT              & 51A DSHOT                \\ \hline
Battery           & 4S 7200mAh & 4S 7200mAh   \\ \hline
Propeller model  & 5045          & T-Motor P13*4.4 \\ \hline
Blade Length     & 127mm         & 165mm           \\ \hline
Number of blades     & 3             & 2               \\ \hline
Total weight     & $\approx 1100g$              & $\approx 2900g$                \\ \hline
Thrust-to-Weight ratio     & 34.98N/kg              & 22.48N/kg                \\ \hline
\end{tabular}
\label{tab:spec_quadrotor}
\end{table}

In our experiments, we used two sets of rotors. The first set uses a DSMX MT2204 motor with 2300KV and a 5045 triple-blade propeller (usually used in a 250mm quadcopter). The second set uses a dual-blade CFProp 13x4.4-inch propeller with a T-motor MN3510 700KV (traditionally used in a 500mm platform). A more detailed specification of both quadrotors is presented in Table~\ref{tab:spec_quadrotor}. Table~\ref{tab:thrust_comparison} shows the results of the thrust estimation compared to its ground truth value for both 250mm and 500mm rotors, respectively. This experiment shows that, with the coefficients presented in Table~\ref{tab:aerodynamic_coefficients}, the method overestimates the thrust for the 250mm and underestimates the thrust for the 500mm, and in both cases, the Pearson correlation between the estimated thrust and the ground truth are $0.997$ and $0.995$ respectively. This indicates that the estimated thrust and the actual thrust are linearly correlated and that the aerodynamic parameters listed in Table~\ref{tab:aerodynamic_coefficients} give a coherent result for static thrust tests for these rotors, and thus, we can state that the thrust estimate is simply a scaled version of the ground truth thrust.

The PX4 control pipeline represents the thrust setpoints of each rotor, calculated by the Mixer block, as relative values of thrust in the interval $[0, 1]$, where 0 is minimum thrust, and 1 is maximum thrust. Thus, using this thrust estimation approach in a real UAV requires no more than performing a calibration experiment in flight, inputting the maximum throttle, and storing the maximum estimated thrust for each rotor. This maximum thrust will be converted into the relative thrust that will be fed back into the thrust controller. The only parameters to set for the thrust estimation are the mass and radius of the propeller.

\begin{table}[]
\centering
\caption{Comparison between estimated $\hat{T}$ and actual thrust $T$ (both in Newtons) and their correlation for both 250mm and 500mm quadrotor rotors.}
\begin{tabular}{|c|ll|ll|}
\hline
               & \multicolumn{2}{c|}{\textbf{250mm}}                     & \multicolumn{2}{c|}{\textbf{500mm}}                     \\ \hline
\textbf{Throttle}       & \multicolumn{1}{c}{$\hat{\text{T}}$} & \multicolumn{1}{c|}{T} & \multicolumn{1}{c}{$\hat{\text{T}}$} & \multicolumn{1}{c|}{T} \\ \hline
10       & 2.33           & 0.19           & 0.241          & 0.32       \\ \hline
20       & 7.35           & 0.63            & 0.83          & 1.26        \\ \hline
30       & 14.29           & 1.23             & 1.71           & 2.57        \\ \hline
40       & 22.58           & 2.06            & 2.86           & 4.55        \\ \hline
50       & 32.54            & 2.97            & 3.91           & 5.92        \\ \hline
60       & 43.78     & 4.25      & 5.93     & 8.94  \\ \hline
70       & 56.55    & 5.45      & 7.08     & 11.02 \\ \hline
80       & 69.62       & 7.39      & 8.36     & 13.96 \\ \hline
90       & 79.26     & 8.63      & 9.17     & 15.63 \\ \hline
100      & 85.71     & 9.62      & 9.56     & 16.30  \\ \hline
\textbf{Pearson Corr.} & \multicolumn{2}{c|}{$\bm{0.997}$}                    & \multicolumn{2}{c|}{$\bm{0.995}$}                  \\ \hline
\end{tabular}
\label{tab:thrust_comparison}
\end{table}

The algorithms and techniques presented in this paper were implemented in the PX4 environment. PX4 is a highly portable autopilot with a flight control solution for a diverse range of UAVs, such as fixed wings, VTOL, and others. The implementation of our approach in the PX4 code is also available as an open-source\footnote{\url{https://github.com/LASER-Robotics/px4_firmware/tree/thrust_control_1.13.2}}. Furthermore, the rotor parameters were initialized with the aerodynamic parameters shown in Table~\ref{tab:aerodynamic_coefficients}, and with $N = 20$, $\Delta = 10^{-1}$, and $\epsilon = 10^{-5}$. 
We first implemented and tested the proposed algorithms in the mrs\_uav simulator~\cite{baca2021mrs}, a Gazebo ROS-based quadrotor simulator that has a digital twin of the 500mm (x500) UAV. This simulator runs PX4 as software-in-the-loop, allowing us to implement and debug the code before compiling and loading the firmware into the hardware. This initial test was important for debugging the code, but validation of the algorithm was not possible because the simulator did not provide coherent values for the electromechanical values needed to validate the algorithm.

\subsection{Thrust control}

We use the estimated thrust as the feedback for a closed-loop thrust controller. This controller is an augmented PID with feedforward of the setpoint to increase convergence speed. The control law can then be formulated as

\begin{equation}
    \bm{u} = K_{\text{ff}}\bm{T_{\text{sp}}} + K_p \bm{e_{T}} + K_i \int \bm{e_T} + K_d \frac{d\bm{e_T}}{dt},
    \label{eq:pid_thrust}
\end{equation}

\noindent where $\bm{u} \in \mathbb{R}^{4 \times 1}$ is the vector of rotor speed command for each rotor and $\bm{e_T} \in \mathbb{R}^{4 \times 1}$ is the vector of thrust error, given by $\bm{e_T} = \bm{T}_{sp} - \bm{\hat{T}}$.

\begin{figure}[htbp]
    \centering
    \includegraphics[width=0.5\textwidth, height=0.2\textheight]{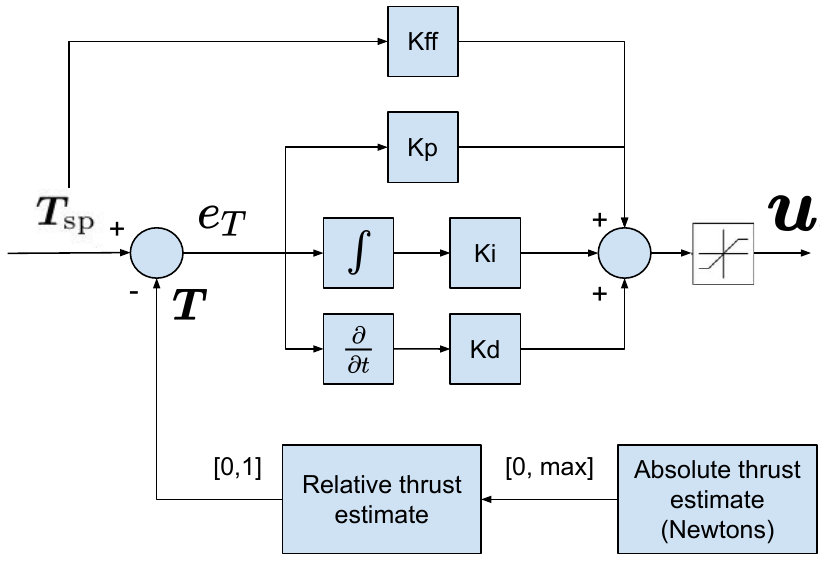}
    \caption{Thrust control diagram.}
    \label{fig:thrust_control_diagram}
\end{figure}

The proposed rotor control at the PX4 level is performed and shown in Fig.~\ref{fig:thrust_control_diagram}. Internally, the controller uses normalized values to represent the thrust commands within the range of $[0, 1]$, where one represents the maximum rotor thrust. To provide feedback on the thrust estimation in the thrust control, we need to convert the absolute thrust value given by the thrust estimation to the relative thrust. This is performed by taking the maximum thrust value during flight by setting the maximum throttle. The resulting value is used as a higher bound for the conversion from absolute to relative thrust.

The tuning of the thrust controller was performed through several flight experiments and observations of the thrust curves. In the following section, we discuss the results using two types of UAVs, the 250mm and the 500mm, where 250mm and 500mm are the distance between propellers in millimeters. For the 250mm UAV, the controller gains are $K_p = 1.0, K_i = 0.3, K_d = 0, K_{ff} = 0.8$, and for the 500mm UAV the gains are $K_p = 2.15, K_i = 0.5, K_d = 0, K_{ff} = 0.9$. 

\section{Results}

We verified the proposed methods with several outdoor experiments, consisting of autonomously following a rectangular path defined by four waypoints and then calculating the error between the desired and actual acceleration, velocity, position, and thrust. This path was defined to be perpendicular to the average wind direction to test the lateral robustness of both approaches. The video of the experiment is also available\footnote{\url{https://youtu.be/FTjXzYDOK5M}}. The idea was to validate the thrust control by the statistical analysis of the error distribution over 32 outdoor experiments for each quadrotor, comparing its robustness to the standard thrust-to-speed approach under varying wind conditions.

\begin{figure}[htbp]
    \centering    \includegraphics[width=0.45\textwidth, height=0.25\textheight]{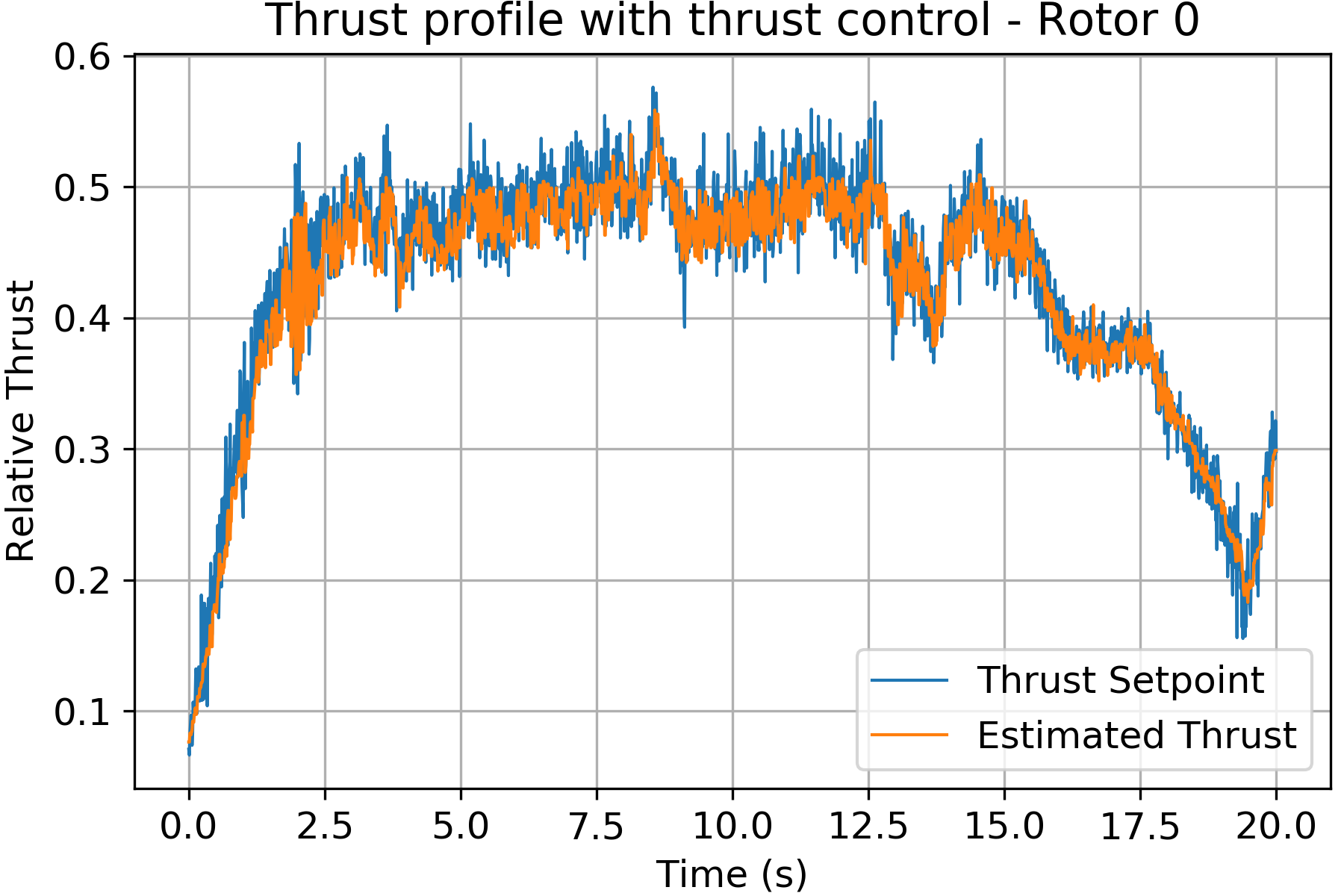}
    \caption{Thrust setpoint tracking with f-PID control.}
    \label{fig:thrust_control_pid}
\end{figure}

The specifications of the quadrotors used in the experiments are shown in Table~\ref{tab:spec_quadrotor}, and we used the standard navigation parameters for the PX4, except for the maximum acceleration, which was set to $4\text{m/s}^2$. To compare against the thrust control, we used the standard approach for the thrust-to-speed map (Equation~\ref{eq:pam_simple}), adding a linear parameter to account for battery depletion, similar to the work of Hentzen et al.\cite{hentzen2019disturbance}. The tests were performed alternately, and for each run with the same quadrotor, the thrust-to-speed map and the thrust control were switched. This experiment methodology aims to test multiple initial voltage conditions to reduce the bias from battery depletion between the approaches. In all experiments, we used 7200mAh 4S Li-Po batteries with an initial voltage of 16.8V when fully charged. The same experimental methodology was used for both the 250mm UAV and the larger 500mm UAV. We performed 8 flights with each battery, with a total of 16 trials. The maximum speed reached by the 250mm and 500mm UAV was $11.3$m/s and $12.7$m/s, respectively. For the experiment with both quadrotors, the wind direction was NW, which is perpendicular to the path shown in the video, but during the trials with the 250mm the average wind speed was 3.05m/s and for the 500mm the wind speed was significantly higher at 4.8m/s. Fig.~\ref{fig:thrust_control_pid} shows the result of one experiment run where we compare the desired thrust with the estimated one in a closed loop (the controlled and estimated thrust).

\begin{figure*}[h!]
    \centering
    \begin{subfigure}[t]{0.5\textwidth}
        \centering    
        \includegraphics[width=\textwidth, height=0.2\textheight]{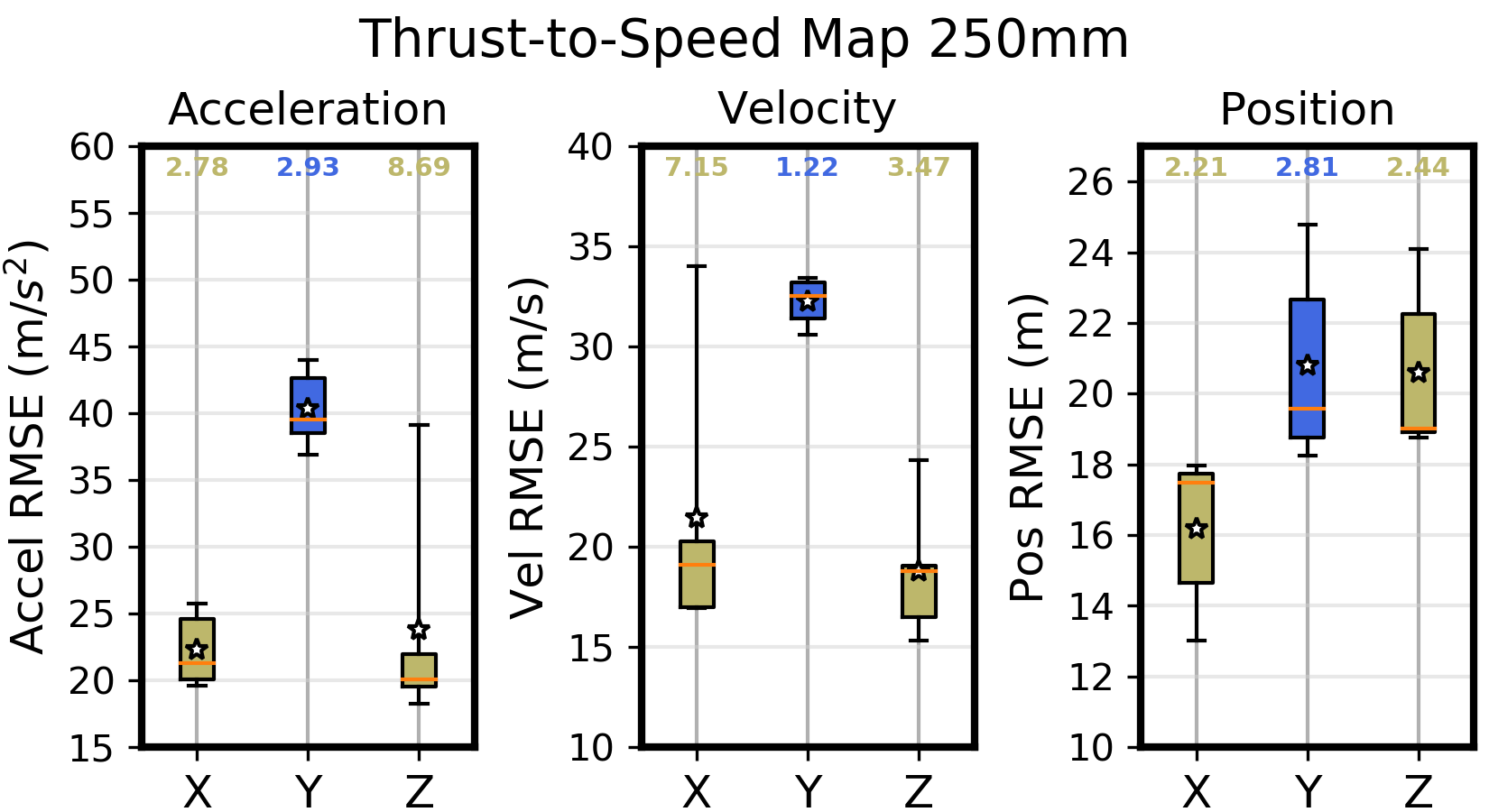}
        \caption{Results with the thrust-to-speed map (16 in total).}
        \label{fig:simple_map_rmse_250}
    \end{subfigure}%
    ~ 
    \begin{subfigure}[t]{0.5\textwidth}
         \centering    
         \includegraphics[width=\textwidth, height=0.2\textheight]{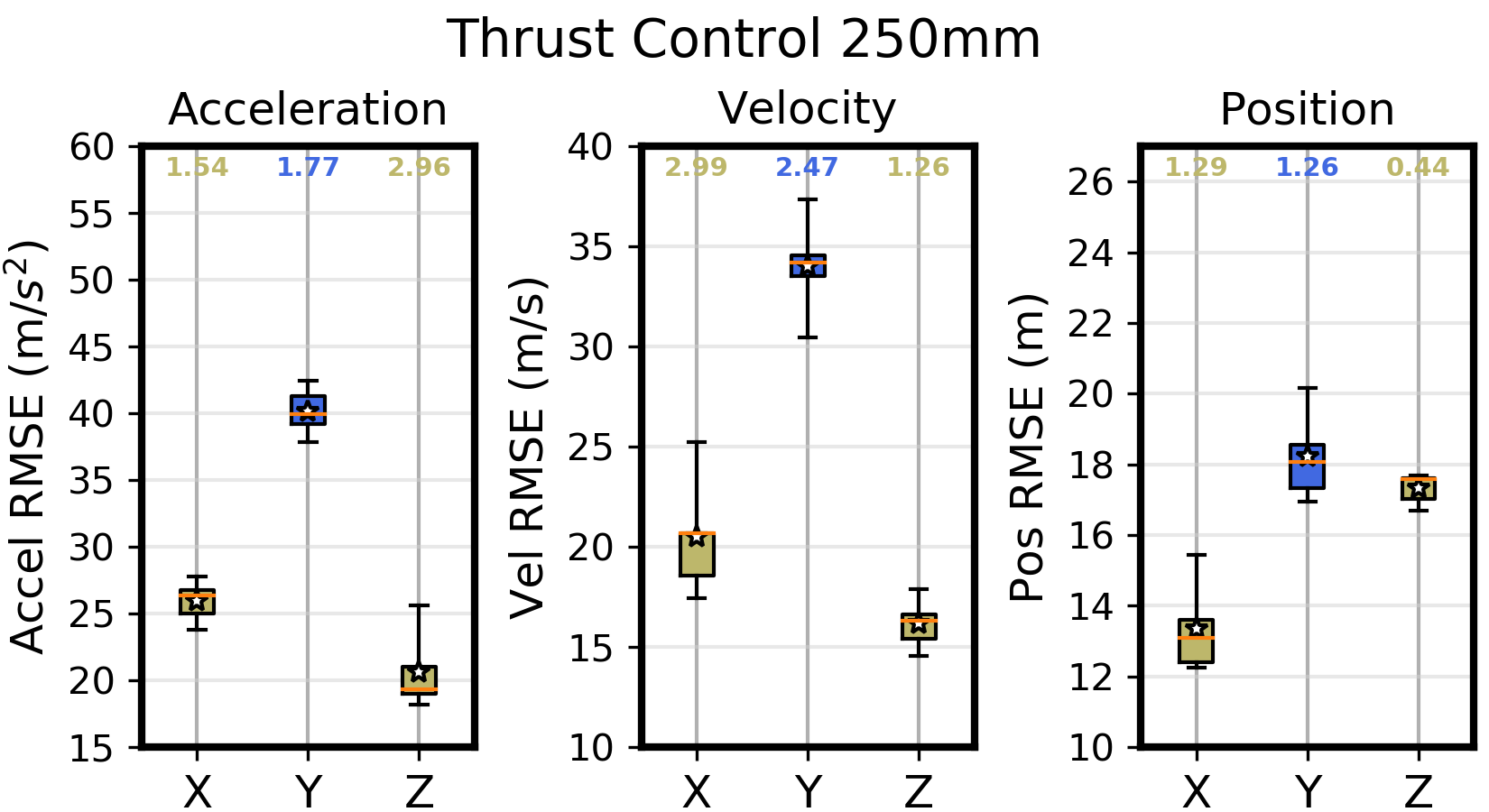}
        \caption{Results with the thrust control (16 in total).}
        \label{fig:thrust_control_rmse_250}
    \end{subfigure}%
    \caption{Acceleration (x, y, z), Velocity (x, y, z), and Position (x, y, z) RMSE distribution for the 32 experiments with the 250mm quadrotor using the standard thrust-to-speed map (16 trials) and the thrust control (16 trials). For all boxplots the star shows the median, the orange line represents the mean value and the numbers at the top are the standard deviation.}
\end{figure*}

\begin{figure*}[h!]
    \centering
    \begin{subfigure}[t]{0.5\textwidth}
        \centering    
        \includegraphics[width=\textwidth, height=0.2\textheight]{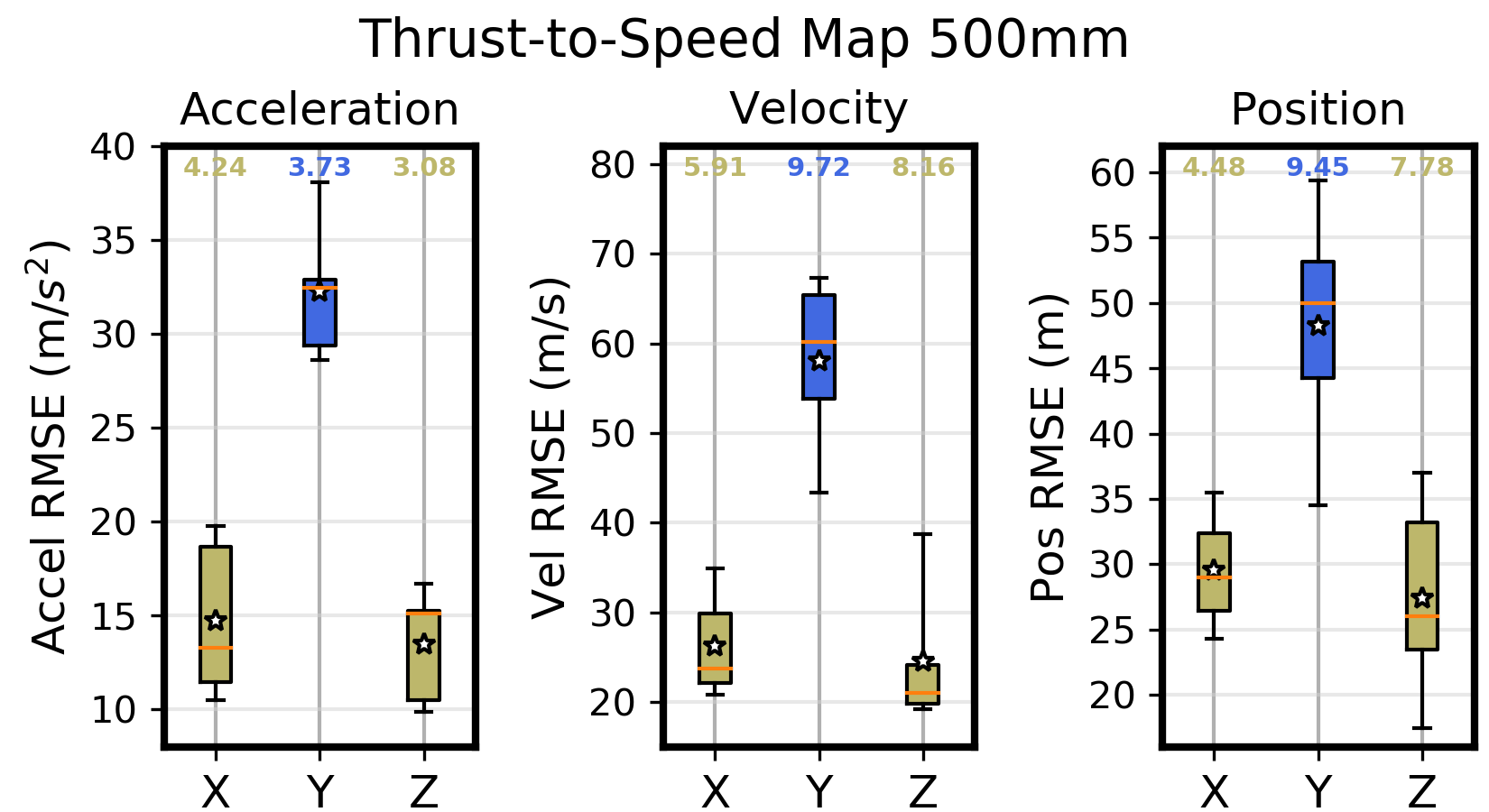}
        \caption{Results with the thrust-to-speed map (16 in total).}
        \label{fig:simple_map_rmse_500}
    \end{subfigure}%
    ~ 
    \begin{subfigure}[t]{0.5\textwidth}
        \centering    
        \includegraphics[width=\textwidth, height=0.2\textheight]{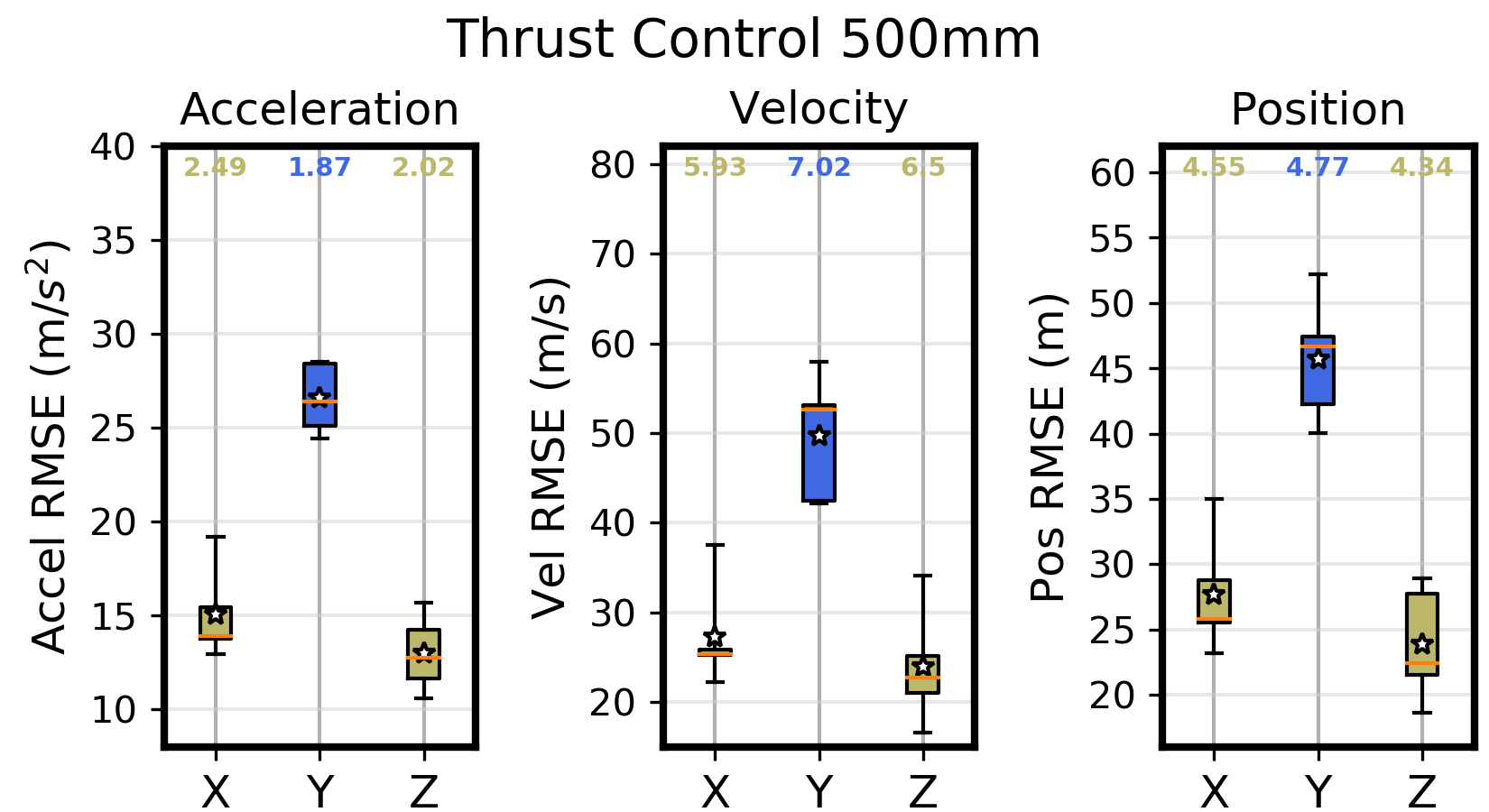}
        \caption{Results with the thrust control (16 in total).}
        \label{fig:thrust_control_rmse_500}
    \end{subfigure}
    \caption{Acceleration, Velocity, and Position RMSE distribution for the 32 experiments with the 500mm quadrotor using the standard thrust-to-speed map (16 trials) and the thrust control (16 trials).}
\end{figure*}

\begin{figure*}[h!]
    \centering
    \begin{subfigure}[t]{0.5\textwidth}
         \centering    
        \includegraphics[width=\textwidth, height=0.2\textheight]{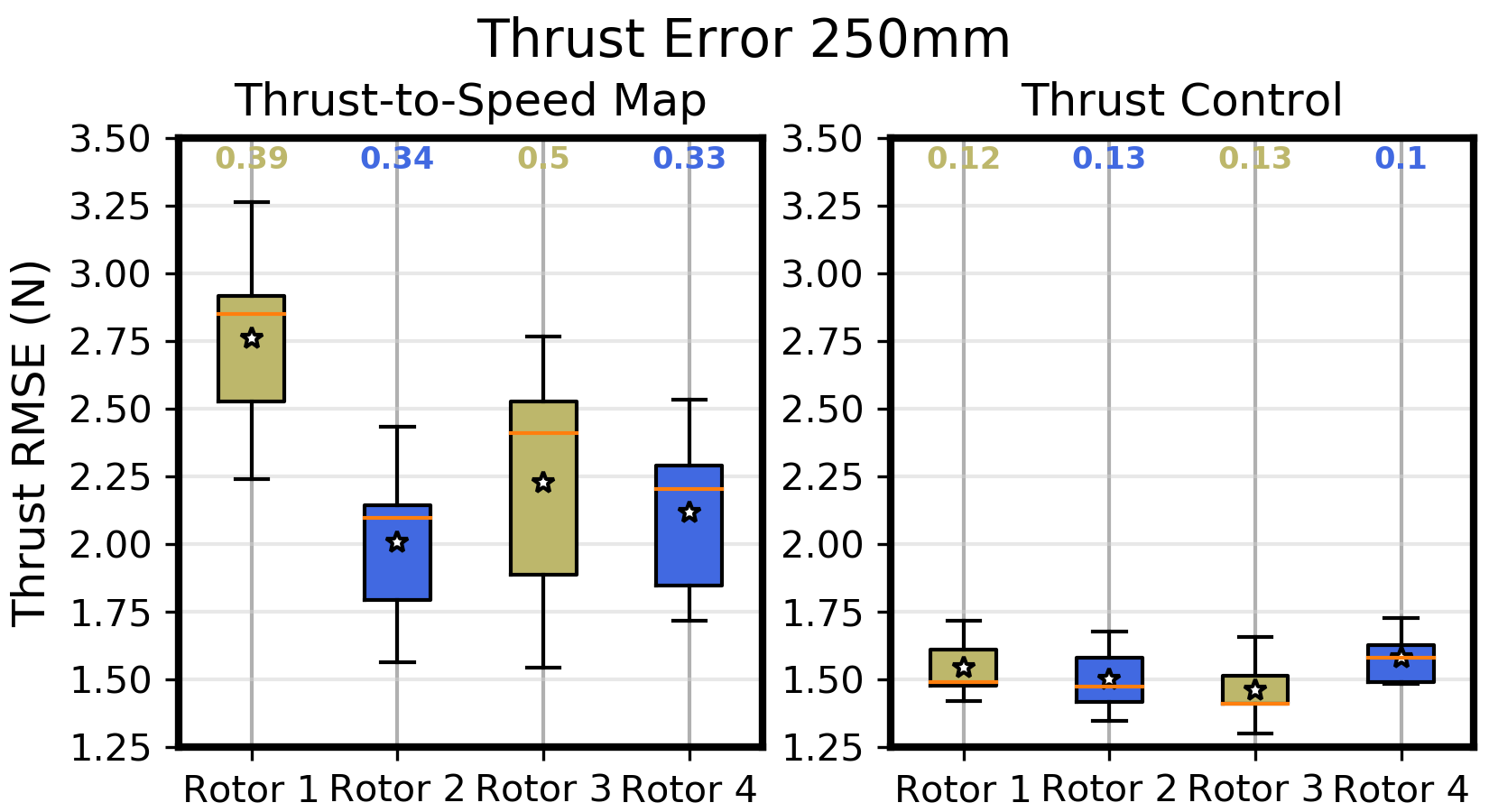}
        \caption{Results for the 250mm quadrotor (16 in total for each method).}
        \label{fig:thrust_rmse_250}
    \end{subfigure}%
    ~ 
    \begin{subfigure}[t]{0.5\textwidth}
        \centering    
        \includegraphics[width=\textwidth, height=0.2\textheight]{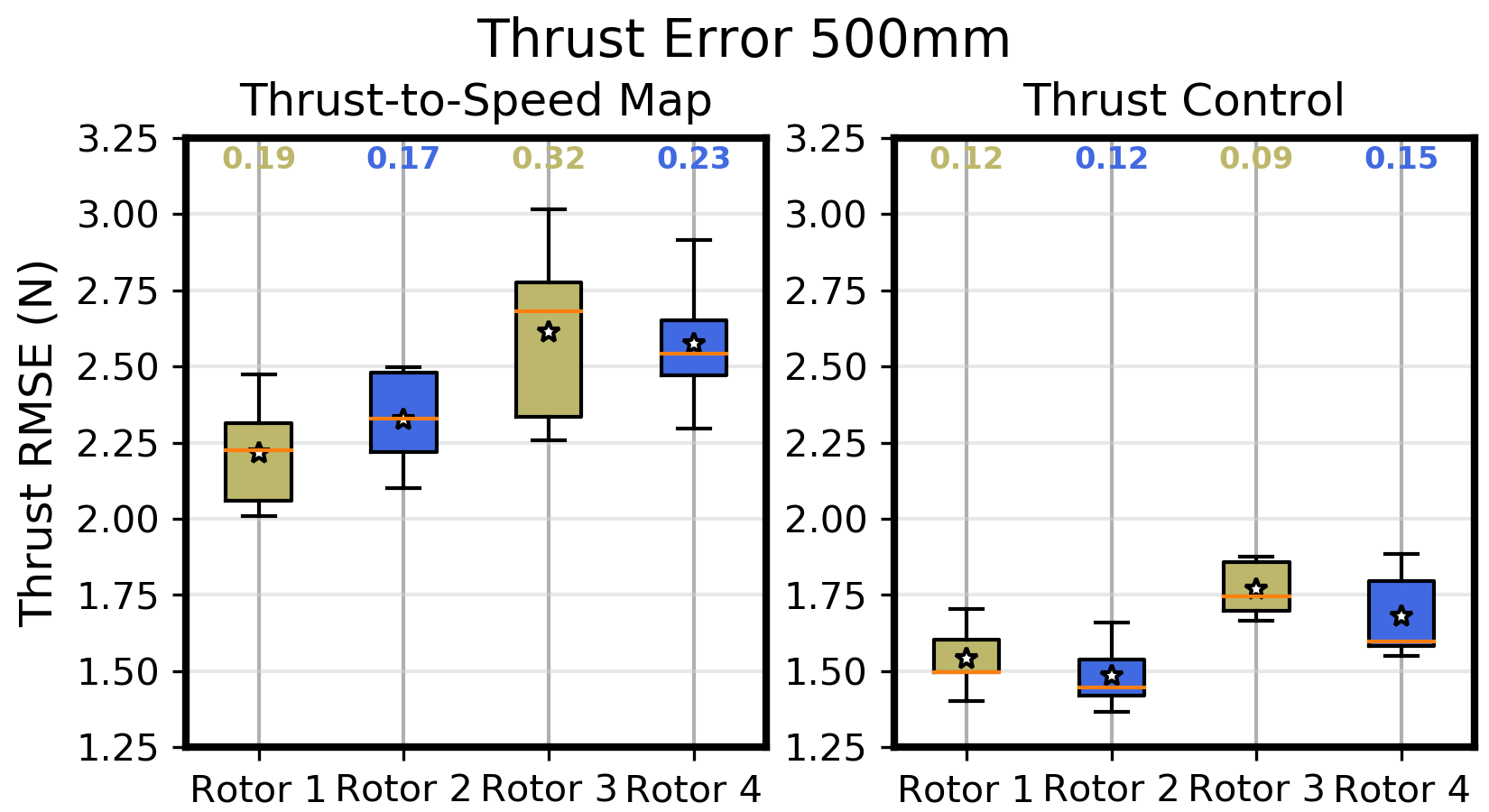}
        \caption{Results for the 500mm quadrotor (16 in total for each method).}
        \label{fig:thrust_rmse_500}
    \end{subfigure}
    \caption{Thrust RMSE distribution using the thrust-to-speed map and thrust estimation and control for the 32 experiments with the 250mm and 500mm each.}
\end{figure*}

To evaluate the results, we analyze the tracking error for the acceleration, velocity, position, and rotor thrust of all runs, using the root-mean-square (RMSE) metric. The RMSE was found by comparing the acceleration, velocity, position, and thrust references set by the high-level navigation system with the actual trajectory given by the internal estimators. In the end, we had a set of 32 RMSEs for each quadrotor, with 16 RMSEs for each method, and a statistical analysis between the errors of the thrust control and the thrust-to-speed map was performed. Figs.~\ref{fig:simple_map_rmse_250}, \ref{fig:thrust_control_rmse_250}, and \ref{fig:thrust_rmse_250} show the results for the 250mm UAV, while Figs.~\ref{fig:simple_map_rmse_500}, \ref{fig:thrust_control_rmse_500}, and \ref{fig:thrust_rmse_500} present the results for the 500mm UAV.

A noticeable trend in the acceleration, velocity, and position results is that for both platforms, the mean, median, and standard deviation for the thrust control trials are smaller than for those using the thrust-to-speed map. This indicates that the thrust control was more consistent between runs, indicating a higher robustness to aerodynamic effects. The only exceptions are the values of mean and median of the acceleration in x and y of the RMSE for the 250mm UAV, in which the thrust-to-speed map slightly outperforms the thrust control. This can be explained by the fact that the rotors did not exceed more than $0.7$ of relative thrust in all experiments, maintaining a mean of $\approx 0.5$ and $\approx 0.4$ for the 250mm and 500mm quadrotors, respectively. The average speed set for the experiments affects both UAVs differently due to their size, air resistance, and thrust-to-weight ratio. However, in a more aggressive maneuvering scenario, this result would not appear, suggesting that higher speed maneuvers may be desirable to thoroughly test the capabilities of a smaller aggressive UAV, such as the 250mm. Another noticeable trend in all results is the higher error in Y due to the drift caused by the lateral wind. This is more apparent in the 500mm UAV results. Still, the thrust control performed better in both the average error and the standard deviation.

By analyzing the thrust error plots, we can note again that the thrust control outperforms the thrust-to-speed map in both standard deviation and mean performance. This is to be expected since, as shown in Fig.~\ref{fig:thrust_control_pid}, the thrust control acts so that the thrust estimate closely follows the thrust setpoint. There is an apparent discrepancy between the desired thrust and the thrust estimate at the rotors for the thrust-to-speed map approach. In this case, the thrust setpoint does not necessarily represent the actual thrust required by the rotors but rather the motor speed that results in the desired motion, which the high-level controllers compensate for. Thus, despite the thrust outputs, the actual control variable in this case is the motor speed.

An analysis of the computational load shows that the thrust estimation takes on average 25.23$\mu$s, with a minimum of 12$\mu$s and a maximum of 52$\mu$s. This variation is due to the convergence of algorithm 1, which can take between 5 and 15 iterations. With this runtime, the thrust estimation module can safely run at 500Hz during flight. The thrust control also adds a computational load to the Pixhawk, with an average run time of 5.29$\mu$s (minimum 4.0$\mu$s and maximum 12.0$\mu$s), while the thrust-to-speed approach takes an average of 1.85$\mu$s (minimum 1.0$\mu$s and maximum 9.0$\mu$s). In both cases, there's no noticeable loss of performance on the Pixhawk 6C, and the thrust control module runs at 500Hz during flight.

A noticeable drawback of the proposed thrust control is the significantly increased power consumption due to the increased rotor commands, as shown in Figure~\ref{fig:thrust_control_pid}. During the experiments with the 250mm, the average power consumption with thrust-to-speed was 416.3mAh, while with thrust control, it was 491.5mAh, an increase of 15.30\%. This trend is repeated for the 500mm UAV, with 653.2mAh for thrust-to-speed and 822.7mAh for thrust control, an increase of 20.60\%. The overall increase is caused by the more aggressive rotor commands introduced by the thrust control, and the main factors causing high-frequency noise in rotor control are the wind conditions and the high gains of the controller, as can be seen from the 500mm results. This indicates that the proposed thrust control is more suitable for applications where precise control of the quadrotor motion is required and power consumption constraints are negligible or in extreme weather situations with high-speed winds to improve the safety and maneuverability of the UAV.

\section{Conclusion}

This paper presents a generalized thrust estimation and control approach for multirotor UAVs implemented at the PX4 framework level. The proposed approach requires minimal calibration and has been validated through several outdoor experiments using two different quadrotors with different rotors. The statistical analysis of several outdoor experiments shows an improvement in the robustness of the thrust control to wind gusts compared to the standard thrust-to-speed rotor map for both platforms used. Despite the increased computational load from the proposed methods, the estimator and the controller are executed in real-time on a Pixhawk 6C, allowing onboard computation of the thrust estimate and accurate thrust control of the rotors. Possible future work consists of running the experiments with a wide range of platforms and comparing the proposed thrust control methodology to other techniques to further strengthen the generability, and also improve the thrust estimation by removing the assumptions used to decouple the horizontal force (see equation~(\ref{eq:ch_bemt2})) and calculating the $\mu$ and $C_H$ directly from the acceleration given by the IMU measurements, which are readily available in the autopilot board.

\ifCLASSOPTIONcaptionsoff
  \newpage
\fi

\bibliographystyle{IEEEtran}
\bibliography{preprint}

\end{document}